\documentclass[11pt, a4paper]{article}
\usepackage[margin=1in]{geometry}
\usepackage{setspace}
\usepackage[T1]{fontenc}
\usepackage[utf8]{inputenc}
\usepackage{mathpazo}            
\usepackage{inconsolata}         
\usepackage{microtype}           
\usepackage{newunicodechar}
\newunicodechar{─}{-}

\usepackage{amsmath,amssymb}

\usepackage{graphicx}
\usepackage{fancyhdr}
\usepackage{booktabs}
\usepackage{caption}
\usepackage{float}
\usepackage{natbib}

\usepackage[colorlinks=true,citecolor=blue,linkcolor=black,urlcolor=blue]{hyperref}

\usepackage{enumitem}
\setlist{nosep, leftmargin=*}

\usepackage{titlesec}
\titleformat{\section}{\large\bfseries}{\thesection}{0.8em}{}
\titleformat{\subsection}{\normalsize\bfseries}{\thesubsection}{0.6em}{}
\titlespacing*{\section}{0pt}{1.4ex plus .4ex minus .2ex}{0.8ex plus .2ex}
\titlespacing*{\subsection}{0pt}{1.0ex plus .3ex minus .2ex}{0.5ex plus .1ex}

\newcommand{\accbase}{\text{Acc}_{\text{base}}}
\newcommand{\accrob}{\text{Acc}_{\text{rob}}}
\newcommand{\btr}{\text{BTR}}

\title{\textbf{MamaBench: Benchmarking LLM Robustness in Maternal\\[2pt]
and Child Health Diagnosis through Counterfactual\\[2pt]
Clinical Perturbation}}

\usepackage{authblk}

\author[1,2]{Thanni Adewuyi\thanks{\texttt{tadewuyi949@stu.ui.edu.ng}}}
\author[1]{Anuoluwa Sotome}
\author[1]{Samuel Okoko}
\author[1]{Angel Ezendu}
\author[1]{Oluwafunke Akinbuwa}
\author[1]{Oluwaseun Odunsi}
\author[1]{Oluwasegun Oguntuase}
\author[1]{Ifeoma Nwabueze}
\author[1]{Abiodun Adereni\thanks{Corresponding author. \texttt{biodun@helpmum.org}}}

\affil[1]{Helpmum Africa}
\affil[2]{University of Ibadan}

\date{}

\begin{document}
\maketitle
\thispagestyle{empty}

\begin{abstract}
\noindent
Large language models achieve strong scores on medical benchmarks, yet these benchmarks evaluate each question in isolation, providing no measure of whether a system can distinguish clinically similar presentations requiring different interventions. We introduce \textbf{MamaBench}, the first counterfactual benchmark for maternal and paediatric AI: 434 expert-authored clinical narratives in 217 pairs across 371 pathologies, evaluated via the \textbf{Bias Trap Rate (BTR)}, the conditional probability that a model fails the counterfactual given success on the base case. We propose \textbf{Evidence-Anchored RAG (EA-RAG)}, a three-stage retrieval method that replaces aggregate similarity with an evidence coverage objective through clinical parameter extraction, coverage auditing, and contrastive sub-queries. Across eight configurations of four frontier LLMs, base accuracy overstates robust accuracy by 16-28 percentage points in every model. EA-RAG achieves 20.3\% BTR and 65.0\% robust accuracy on Claude Sonnet 4.6, a 5.5 percentage point BTR reduction without degrading base accuracy. The residual ~20\% BTR confirms that counterfactual robustness in clinical AI remains an open challenge.
\end{abstract}

\vspace{0.3em}
\vspace{0.8em}

\section{Introduction}

LLMs now pass medical licensing exams \citep{singhal2023large,nori2023capabilities} and outperform physician baselines on clinical QA \citep{jin2021disease,pal2022medmcqa}, accelerating deployment via retrieval-augmented generation for domain-grounded decision support \citep{lewis2020retrieval,zakka2024almanac}. In maternal healthcare where a diagnostic error threatens two lives, such systems carry both exceptional promise and exceptional risk \citep{topol2019high}.

Yet current benchmarks (MedQA, PubMedQA, MedMCQA) score each question independently, hiding a dangerous failure mode: a system may answer correctly when the presentation matches its prior, yet fail when a single discriminative parameter shifts the correct diagnosis. Contrast sets \citep{gardner2020evaluating} and counterfactual augmentation \citep{kaushik2020learning} exposed this fragility in general NLP; MedEinst \citep{chen2026medeinst} extended it to medical LLMs via cases derived from DDXPlus \citep{fansitchango2022ddxplus}. No counterfactual benchmark yet exists for maternal and paediatric care the domain where clinical stakes are highest.

We close this gap with two contributions:

\begin{enumerate}
  \item \textbf{MamaBench}: 434 expert-authored clinical narratives in 217 counterfactual pairs across 371 pathologies, written in first-person patient-reported format by a three-member clinical team. We formalise \textit{Diagnostic Fixation} (a model gets the base case right but rigidly persists when discriminative parameters shift) and adopt the \textbf{Bias Trap Rate (BTR)} \citep{chen2026medeinst} to quantify it.

  \item \textbf{Evidence-Anchored RAG (EA-RAG)}: A three-stage inference-time pipeline that replaces aggregate similarity retrieval with an \textbf{evidence coverage} objective: extracting typed clinical parameters, auditing retrieval coverage, and filling gaps via contrastive sub-queries, guided by a taxonomy-grounded reasoning scaffold.
\end{enumerate}

Across eight configurations of four frontier LLMs we find: (i) base accuracy overstates robust accuracy by 16--28\,percentage point in \textit{every} model; (ii) vanilla RAG provides no counterfactual benefit; (iii) EA-RAG reduces BTR by 5.5\,percentage point on the strongest model without degrading base accuracy; (iv) even the best system still fails one in five counterfactual pairs.

\section{Related Work}

\paragraph{Medical QA benchmarks.}
MedQA \citep{jin2021disease}, PubMedQA \citep{jin2019pubmedqa}, and MedMCQA \citep{pal2022medmcqa} evaluate isolated questions, enabling landmark results from Med-PaLM \citep{singhal2023large} and GPT-4 \citep{nori2023capabilities} but providing no pair-level discrimination signal.

\paragraph{Counterfactual and robustness evaluation.}
Contrast sets \citep{gardner2020evaluating} and counterfactually augmented data \citep{kaushik2020learning} showed that high-accuracy models exploit surface cues; CheckList \citep{ribeiro2020beyond} formalised behavioural testing via structured perturbations. MedEinst \citep{chen2026medeinst} benchmarks the Einstellung effect \citep{luchins1942mechanization} through counterfactual diagnosis on DDXPlus \citep{fansitchango2022ddxplus}, while \citet{li2025benchmarking} evaluated clinical RAG robustness using a specialized counterfactual retrieval testbed. MamaBench differs in being (a)~entirely expert-authored, (b)~focused on maternal/paediatric care, and (c)~paired with a retrieval-level intervention. Complementary robustness studies address paraphrasing and demographic perturbations \citep{pfohl2024toolbox} but not diagnostic fixation.

\paragraph{Retrieval-augmented generation.}
RAG \citep{lewis2020retrieval} has been refined by dense retrieval \citep{karpukhin2020dense}, adaptive strategies (FLARE, \citealt{jiang2023active}; Self-RAG, \citealt{asai2024selfrag}; IRCoT, \citealt{trivedi2023interleaving}), clinical example-selection frameworks like MMRAG \citep{zhan2025mmrag}, and clinical deployment \citep{zakka2024almanac}. None addresses \textit{retrieval stagnation}: near-identical embeddings for base--counterfactual pairs yield the same retrieved context \citep{shi2023large}. EA-RAG fills this gap through evidence coverage, drawing on query decomposition \citep{khattab2023demonstrate,wang2023query2doc} while operating entirely at inference time.

\paragraph{Structured reasoning.}
CoT \citep{wei2022chain} and Tree of Thoughts \citep{yao2023tree} improve reasoning but are task-agnostic. EA-RAG's scaffold derives from empirical failure taxonomy, targeting specific failure categories (missed critical info, underthinking, factual inaccuracy, overthinking) rather than providing generic instructions.

\section{MamaBench Benchmark}

\subsection{Problem Formulation}

We formalise diagnosis as $f\!:\!\mathcal{X}\!\to\!\mathcal{Y}$, where $\mathcal{X}$ is the space of patient-reported narratives and $\mathcal{Y}$ the label space of 371 pathologies. A \textbf{Counterfactual Pair} $(x_b,x_c)$ comprises a base case $x_b$ with ground truth $y_b$ and a counterfactual $x_c$---a minimal perturbation shifting the ground truth to $y_c\!\neq\!y_b$ while preserving shared context.

\begin{quote}
\textbf{Definition 1 (Diagnostic Fixation).}\; Model $f$ exhibits Diagnostic Fixation on $(x_b,x_c)$ iff:
\begin{equation}
  f(x_b)=y_b \;\wedge\; f(x_c)=y_b \;\wedge\; y_c\neq y_b
\end{equation}
\end{quote}
\vspace{-0.5em}
\noindent The model succeeds in the base case, but persists rigidly when discriminative parameters change analogously to the Einstellung effect \citep{luchins1942mechanization}. Diagnostic Fixation is architecture-agnostic: it can arise from memorised priors, retrieval stagnation, or heuristic shortcuts.

\subsection{Construction and Statistics}

A three-member medical team (obstetrics, gynaecology, paediatrics) authored each narrative in \textbf{first-person patient-reported format}: e.g., \textit{``My baby was delivered at 37 weeks by a traditional birth attendant. He is 10 days old and refusing to eat\ldots''} introducing naturalistic ambiguity absent from structured vignettes \citep{jin2021disease}. Unlike benchmarks derived from existing datasets \citep{fansitchango2022ddxplus} or LLM pipelines \citep{chen2026medeinst}, every case is individually expert-authored.

For each base case, the team constructs a counterfactual by perturbing the fewest parameters necessary to shift the diagnosis, following five categories: symptom substitution, severity escalation, risk factor modification, temporal shift, and comorbidity introduction.

MamaBench totals \textbf{217~pairs} (434~cases): paediatrics 254 (58.5\%), obstetrics/gynaecology 178 (41.0\%), unclassified 2 (0.5\%), covering 371~unique pathologies. All cases passed multi-stage clinical review for diagnostic sufficiency, counterfactual validity, and minimal edit properties.

\subsection{Evaluation Metrics}

Each pair is classified as \textbf{PP}~(both correct), \textbf{PF}~(base correct, counterfactual wrong), \textbf{FP}, or \textbf{FF}. We report three metrics:
\begin{itemize}
  \item \textbf{Base Accuracy:}\; $\accbase = (|\text{PP}|+|\text{PF}|)\,/\,N$
  \item \textbf{Bias Trap Rate} (primary metric):
    \begin{equation}
      \btr = \frac{|\text{PF}|}{|\text{PP}|+|\text{PF}|} \times 100\%
    \end{equation}
  \item \textbf{Robust Accuracy:}\; $\accrob = |\text{PP}|\,/\,N$
\end{itemize}
Predictions are assessed by an LLM-as-judge (Claude Opus~4.5, $T\!=\!0$): lenient on nomenclature, strict on clinically significant modifiers that change management \citep{zheng2023judging}.

\section{Evidence-Anchored RAG}

\begin{figure*}[t]
  \centering
  \includegraphics[width=0.85\textwidth, height=0.4\textheight, keepaspectratio]{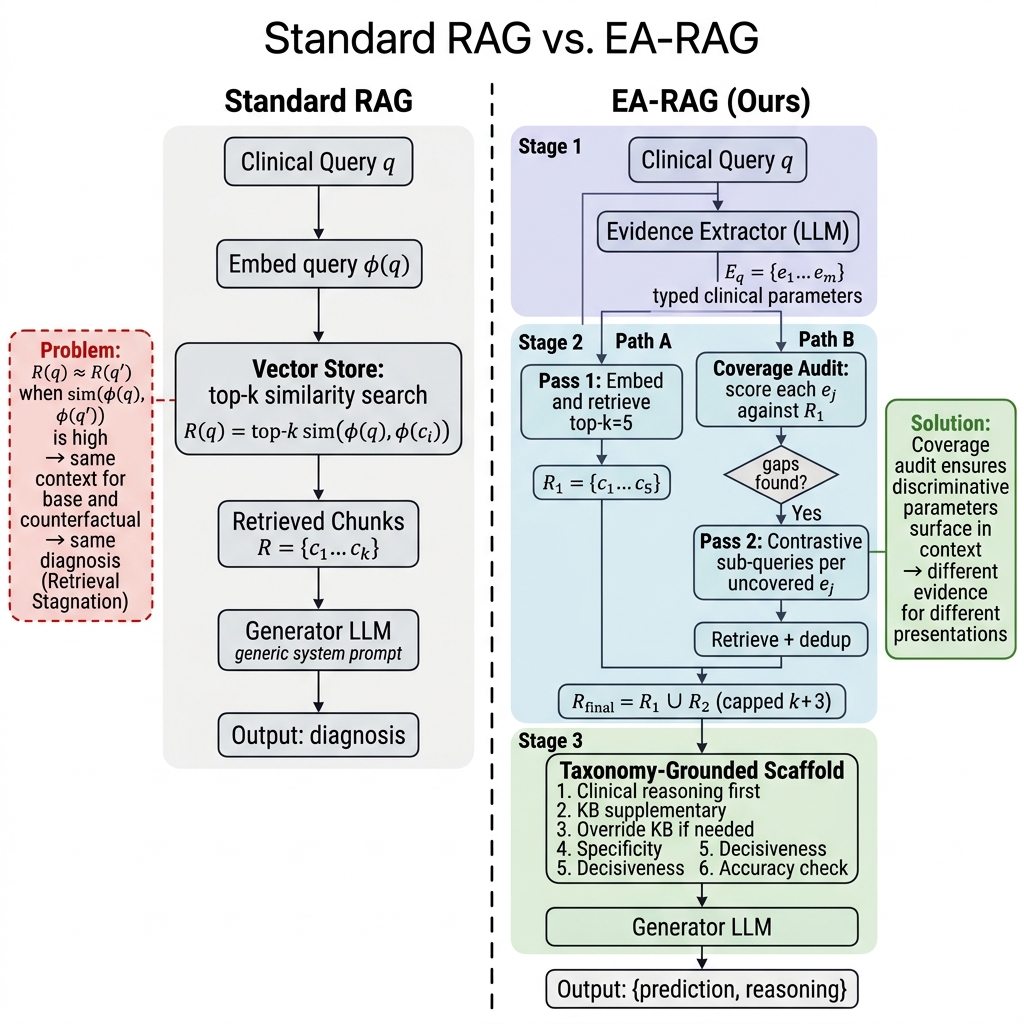}
  \caption{Standard RAG (left) vs.\ EA-RAG (right). Standard RAG retrieves top-$k$ chunks by aggregate query similarity, returning near-identical context for base and counterfactual cases (\textit{retrieval stagnation}). EA-RAG introduces three additional mechanisms: (1)~evidence extraction of typed clinical parameters, (2)~coverage auditing with contrastive sub-queries to fill gaps, and (3)~a taxonomy-grounded generation scaffold.}
  \label{fig:comparison}
\end{figure*}

Standard RAG retrieves the top-$k$ chunks by maximising aggregate query similarity: $R(q)=\text{top-}k\;\text{sim}(\phi(q),\phi(c_i))$. For a base--counterfactual pair $(q,q')$ with $\text{sim}(\phi(q),\phi(q'))\!\approx\!0.94$, this yields $R(q)\!\approx\!R(q')$---the generator reasons from the same context and produces the same diagnosis (\textit{retrieval stagnation}). EA-RAG replaces this with an \textbf{evidence coverage} objective (Fig.~\ref{fig:comparison}).

\subsection{Stage 1: Evidence Extraction}

A structured LLM call (GPT-5.4-nano, $T\!=\!0$) parses the patient narrative into typed clinical parameters $E_q=\{e_1,\ldots,e_m\}$ spanning patient demographics, presenting condition and severity markers, risk factors, contraindications, and management-relevant features. Null fields are informative: their absence signals the query does not specify that parameter. If extraction fails, the system falls back to single-pass retrieval, ensuring EA-RAG never degrades below baseline.

\subsection{Stage 2: Two-Pass Retrieval with Coverage Auditing}

\textbf{Pass~1} retrieves the top-$k$ ($k\!=\!5$) chunks via cosine similarity over \texttt{text-embedding-3-large} embeddings ($d\!=\!3072$). A \textbf{coverage audit} then scores each evidence element $e_j$ against the retrieved set: $\text{cov}(e_j,R^{(1)})=\max_{c_i\in R^{(1)}}\cos(\phi(e_j),\phi(c_i))$. Elements with $\text{cov}<\theta$ ($\theta\!=\!0.6$) form the gap set~$G$.

\textbf{Pass~2} generates a \textit{contrastive sub-query} for each $e_j\!\in\!G$ framed as ``How does $e_j$ change the management of \textit{condition}?'' and retrieves supplementary chunks. After deduplication ($\delta\!=\!0.82$), the final set $R_\text{final}=R^{(1)}\cup R^{(2)}$ is capped at $k\!+\!3\!=\!8$ chunks. This mechanism ensures discriminative parameters surface in the context even when the overall query embedding fails to distinguish them.

\subsection{Stage 3: Taxonomy-Grounded Generation}

The generator receives the narrative, $R_\text{final}$, and a \textbf{scaffold instruction set} derived from error taxonomy analysis of 113 baseline failures:

\begin{table}[H]
\centering\small
\begin{tabular}{@{}lrl@{}}
\toprule
\textbf{Failure Mode} & \textbf{Freq.} & \textbf{Scaffold Response} \\
\midrule
Missed critical info & 48.7\% & Analyse ALL clinical parameters \\
Underthinking        & 31.9\% & Match to specific severity level \\
Factual inaccuracy   & 10.6\% & Check contraindications first \\
Overthinking         &  8.8\% & State conclusion decisively \\
\bottomrule
\end{tabular}
\end{table}

\vspace{-0.5em}
\noindent The scaffold enforces a \textit{clinical-reasoning-first} hierarchy: the model must diagnose from the presentation before consulting retrieved evidence, may override retrieval when clinical reasoning diverges, and must verify the final diagnosis accounts for all key findings. This contrasts with generic CoT prompting \citep{wei2022chain} and training-time approaches like Self-RAG \citep{asai2024selfrag} EA-RAG operates entirely at inference time with no fine-tuning.

\section{Experiments and Results}

\subsection{Setup}

We evaluate eight configurations: five standalone models (GLM-5, Kimi~K2.5, Nemotron~Super~120B, GPT-5.4, Claude~Sonnet~4.6), a vanilla RAG baseline (GPT-4o, $k\!=\!2$), and EA-RAG on GPT-4o ($k\!=\!5$) and Claude~Sonnet~4.6. All at $T\!=\!0$; all 217~pairs evaluated per configuration.

\textbf{Data separation.} To ensure evaluation integrity, no MamaBench case narratives or ground truth labels were included in the RAG or EA-RAG knowledge base. The retrieval corpus consists exclusively of clinical reference material (textbook guidelines, WHO protocols, and clinical management standards); the benchmark data remains unseen by the retrieval system at all times.

\subsection{Main Results}

\begin{table}[t]
\centering\small
\caption{MamaBench results. Best per column in \textbf{bold}. All values are percentages.}
\label{tab:main}
\begin{tabular}{@{}llccc@{}}
\toprule
\textbf{Model} & \textbf{Setup} & $\accbase$ & $\btr\!\downarrow$ & $\accrob\!\uparrow$ \\
\midrule
GLM-5            & Standalone          & 54.8 & 50.4 & 27.2 \\
Kimi K2.5        & Standalone          & 76.0 & 40.0 & 45.6 \\
GPT-4o           & RAG Baseline        & 77.4 & 39.9 & 46.5 \\
Nemotron 120B    & Standalone          & 72.4 & 35.0 & 47.0 \\
GPT-4o           & \textbf{EA-RAG (Ours)}       & 77.0 & 35.9 & 49.3 \\
GPT-5.4          & Standalone          & 77.4 & 25.6 & 57.6 \\
Claude Sonnet 4.6 & Standalone         & \textbf{82.0} & 25.8 & 60.8 \\
Claude Sonnet 4.6 & \textbf{EA-RAG (Ours)} & \ 81.6 & \textbf{20.3} & \textbf{65.0} \\
\bottomrule
\end{tabular}
\end{table}

\begin{figure}[t]
  \centering
  \includegraphics[width=0.85\columnwidth]{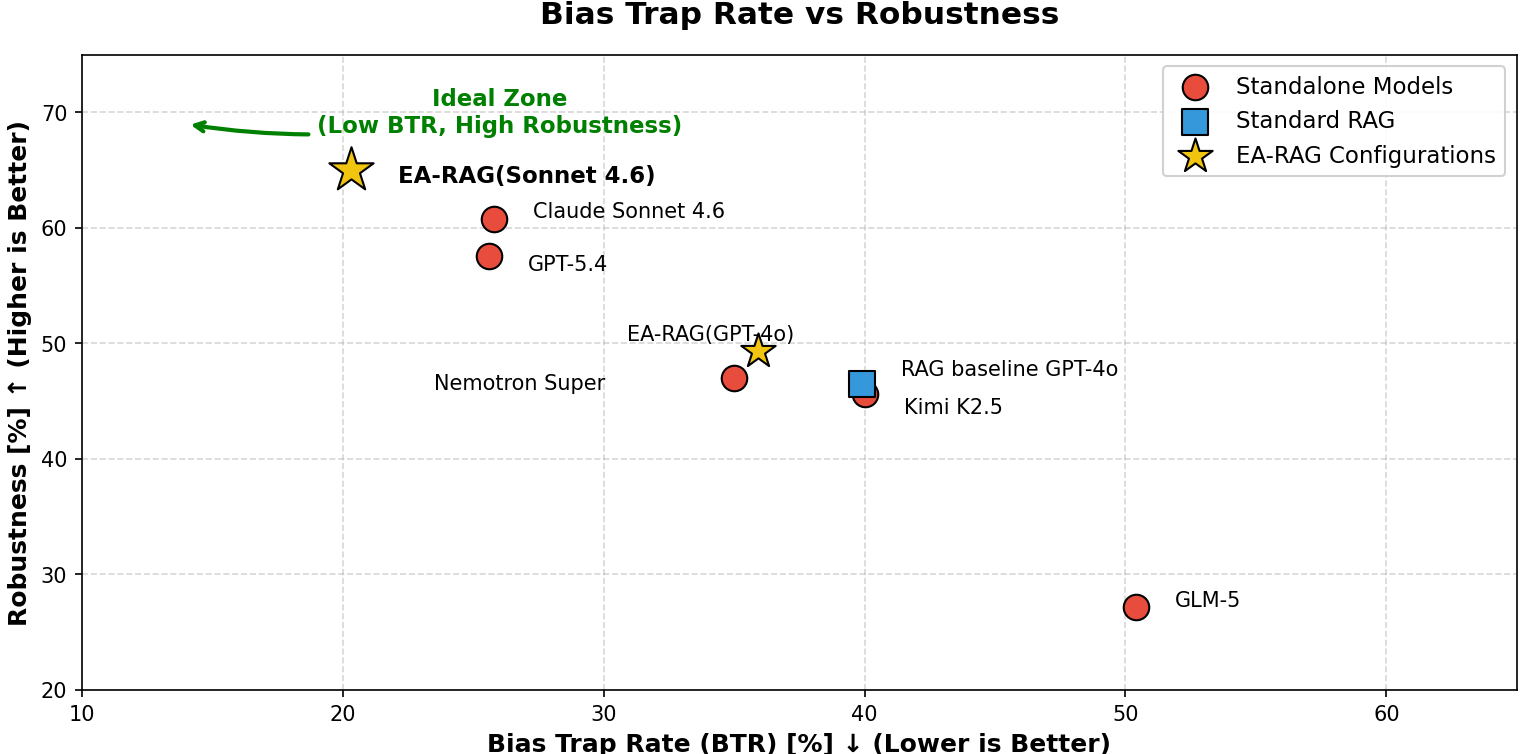}
   \caption{Bias Trap Rate (BTR) vs. Robustness across different model configurations. The x-axis denotes the Bias Trap Rate (where lower percentages are better), and the y-axis represents Robustness (where higher percentages are better). Standalone models are marked as red circles, standard RAG as blue squares, and the proposed EA-RAG configurations as gold stars. The top-left corner represents the ideal performance quadrant. The plot illustrates that while standard RAG improves baseline robustness, the EA-RAG framework uniquely minimizes the bias trap while simultaneously maximizing robustness, driving configurations toward the ideal zone.}
   \label{fig:scatter}
 \end{figure}

\paragraph{Base accuracy is misleading.}
The \textit{robustness gap}---the difference between $\accbase$ and $\accrob$---ranges from 16.2\,pp (Claude~+~EA-RAG: $82.0\!\to\!65.0$) to 27.6\,pp (GLM-5: $54.8\!\to\!27.2$). GPT-4o achieves 77.4\% base accuracy but only 46.5\% robust accuracy ($\btr\!=\!39.9\%$): nearly two in five ``correct'' diagnoses are fragile. No model closes the gap below 16\,pp, confirming counterfactual vulnerability as a systemic property of current clinical AI.

\paragraph{EA-RAG improves robustness without harming base accuracy.}
GPT-4o: $\btr$ drops $39.9\!\to\!35.9\%$ ($-4.0$\,pp); Claude: $25.8\!\to\!20.3\%$ ($-5.5$\,pp). Base accuracy decreases by only 0.4\,pp in both cases. The stronger model benefits more, consistent with evidence coverage being most effective when reasoning capacity is sufficient and retrieval quality is the bottleneck.

\paragraph{Vanilla RAG does not help.}
GPT-4o with RAG ($\btr\!=\!39.9\%$) matches standalone models of similar capability (Kimi~K2.5: 40.0\%). Retrieval stagnation renders standard top-$k$ retrieval invisible to the counterfactual perturbation---confirming that the problem lies in \textit{what} is retrieved, not \textit{whether} retrieval occurs.

\paragraph{Scale reduces but does not eliminate the problem.}
$\btr$ drops from 50.4\% (GLM-5) to 25.6\% (GPT-5.4) across standalone models, but even Claude Sonnet~4.6 standalone (25.8\%) fails on more than one in four pairs where it demonstrates base competence.

\paragraph{The best system still fails 1 in 5.}
Claude~+~EA-RAG achieves the lowest $\btr$ (20.3\%) and highest $\accrob$ (65.0\%). In deployment, this means a system that \textit{appears} reliable will miss discriminative clinical parameters in ${\sim}$20\% of diagnostically ambiguous cases---underscoring that counterfactual robustness remains an open research challenge.

\begin{figure}[t]
   \centering
   \includegraphics[width=0.85\columnwidth]{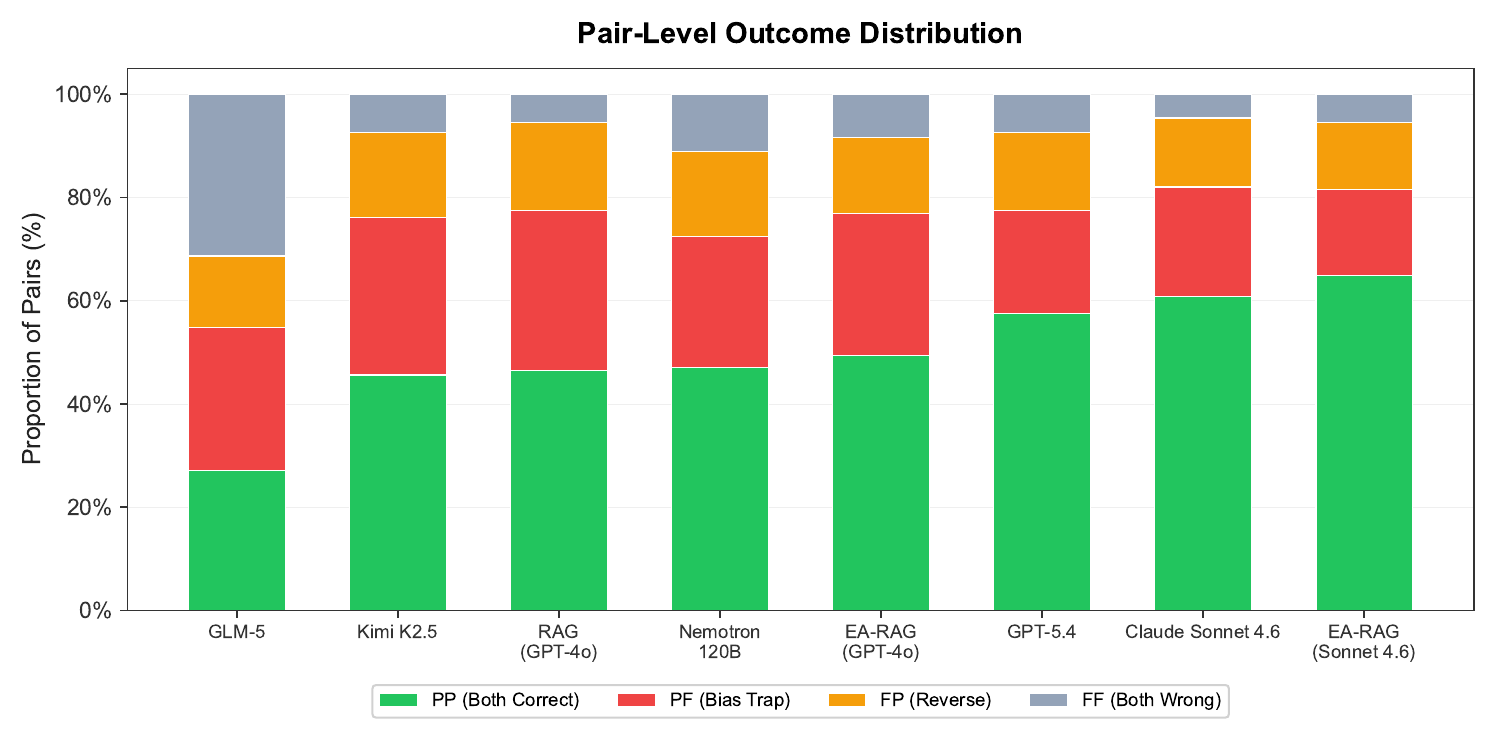}
   \caption{Pair-level outcome distribution. PF (red) = bias traps: model appears competent but fails under perturbation. EA-RAG shrinks PF while expanding PP.}
   \label{fig:stacked}
\end{figure}

\subsection{Ablation Study}

To isolate the contribution of each EA-RAG component, we conduct an additive ablation using GPT-4o as the base model. Starting from the \texttt{k5} configuration (top-$k\!=\!5$, no scaffold, no coverage), we incrementally enable each module (Table~\ref{tab:ablation}).

\begin{table}[t]
\centering\small
\caption{Ablation study on EA-RAG components (GPT-4o). Each row adds one module to the \texttt{k5} baseline. PP/PF/FP/FF are pair-level outcome counts ($N\!=\!217$).}
\label{tab:ablation}
\begin{tabular}{@{}lccccccc@{}}
\toprule
\textbf{Variant} & \textbf{PP} & \textbf{PF} & \textbf{FP} & \textbf{FF} & $\accbase$ & $\btr\!\downarrow$ & $\accrob\!\uparrow$ \\
\midrule
\texttt{k5}            & 94  & 61 & 36 & 26 & 71.4\% & 39.4\% & 43.3\% \\
\texttt{k5\_coverage}  & 94  & 61 & 37 & 25 & 71.4\% & 39.4\% & 43.3\% \\
\texttt{k5\_scaffold}  & 101 & 62 & 35 & 19 & 75.1\% & 38.0\% & 46.5\% \\
\texttt{ea\_rag\_full} & \textbf{107} & \textbf{60} & 32 & 18 & \textbf{77.0\%} & \textbf{35.9\%} & \textbf{49.3\%} \\
\bottomrule
\end{tabular}
\end{table}

\paragraph{Coverage alone is insufficient.}
\texttt{k5\_coverage} adds evidence extraction and coverage auditing but no contrastive sub-queries or scaffold. It produces identical $\btr$ and $\accrob$ to the \texttt{k5} baseline (39.4\% / 43.3\%), indicating that \textit{detecting} coverage gaps without \textit{filling} them or \textit{guiding} generation has negligible effect. The retrieval audit alone does not change what the generator sees.

\paragraph{The scaffold drives the largest single gain.}
\texttt{k5\_scaffold} adds the taxonomy-grounded reasoning scaffold without coverage-aware retrieval. This lifts $\accbase$ by 3.7\,pp (71.4\%\,$\to$\,75.1\%), $\accrob$ by 3.2\,pp, and reduces $\btr$ by 1.4\,pp. Seven additional pairs shift from PF\,$\to$\,PP (and seven from FF\,$\to$\,FP/PP), confirming that structured generation instructions particularly the clinical-reasoning-first hierarchy and permission to override retrieved context---improve diagnostic reasoning even with unchanged retrieval.

\paragraph{The full pipeline is greater than the sum of its parts.}
\texttt{ea\_rag\_full} combines all three stages and achieves the best results on every metric: $\accbase\!=\!77.0\%$, $\btr\!=\!35.9\%$, $\accrob\!=\!49.3\%$. Compared to \texttt{k5\_scaffold}, the addition of coverage-aware retrieval with contrastive sub-queries converts six more pairs from incorrect to correct ($\text{PP}$: 101\,$\to$\,107), yielding a further 2.1\,pp BTR reduction and 2.8\,pp $\accrob$ gain. This confirms that the scaffold and coverage modules are \textit{complementary}: the scaffold tells the generator \textit{how} to reason, while coverage-aware retrieval ensures it has the \textit{evidence} to reason over.

\section{Conclusion}

We introduced MamaBench, the first counterfactual benchmark for maternal and paediatric AI (217 expert-authored pairs, 371 pathologies), and EA-RAG, an evidence-anchored retrieval method. Base accuracy systematically overstates clinical competence by 16--28\,pp across all models. EA-RAG reduces BTR by 5.5\,pp on the strongest model without degrading base accuracy, confirming retrieval quality as the primary bottleneck. The residual 20\% BTR shows retrieval-level interventions alone are insufficient.

\textbf{Limitations.}\; MamaBench's 434 cases cover a single clinical domain in English only, which constrains how far our findings generalize to other specialties or languages. Our reliance on LLM-as-judge evaluation may introduce systematic biases correlated with the judge model's own blind spots \citep{zheng2023judging}, and while expert authoring gives us high-quality counterfactuals, it caps how quickly the benchmark can scale to new cases or domains.

\end{document}